%% file: fove_complete.tex
\newcommand{\exEnd}{$\Box$} 
\newtheorem{theorem}{Theorem} 

\newtheorem{definition}{Definition} 

\documentclass[]{article}

\usepackage{nips12submit_e,times}
\usepackage{amssymb}
 \usepackage{comment}
\usepackage{xcolor}
\usepackage{graphicx}

\title{Lifted Variable Elimination:\\A Novel Operator and Completeness Results}

\author{
Nima Taghipour\quad Daan Fierens\quad Guy Van den Broeck\quad Jesse Davis\quad Hendrik Blockeel\\
Department of Computer Science, KU Leuven\\
Celestijnenlaan 200A, B-3001 Heverlee, Belgium\\
}
\nipsfinalcopy 
\begin{document} 
 
\maketitle 
 
\begin{abstract} 
Various methods for lifted probabilistic inference have been proposed, but our understanding of these methods and the relationships between them is still limited, compared to their propositional counterparts.  The only existing theoretical characterization of lifting is for weighted first-order model counting (WFOMC), which was shown to be complete domain-lifted for the class of 2-logvar models. This paper makes two contributions to lifted variable elimination (LVE).  First, we introduce a novel inference operator called group inversion.  Second, we prove that LVE augmented with this operator is complete in the same sense as WFOMC.
\end{abstract} 
  
\section{Introduction}
\input{Intro}

\section{Representation}
\label{sec:rep}
\input{representation}

\section{(Lifted) Variable Elimination}
\label{sec:c-fove}
\input{c-fove}

\section{A New Operator: Group Inversion}
\label{sec:inversion}
\input{inversion}

\section{Extension for Counting}
\label{sec:extension}
\input{extension}

\section{Completeness Results}
\label{sec:completeness}
\input{completeness}

\section{Conclusion}
\label{sec:discussion}
\input{conclusion}

\section*{Appendix}
\appendix

\input{suppl}

\bibliographystyle{plain} 
\bibliography{LiftedInference}

\end{document}

%% file: Intro.tex
Probabilistic logical models combine graphical models with elements of first-order logic
to compactly model uncertainty in structured
domains (social networks, citation
graphs, etc.)~\cite{DeRaedt2008-PILP,Getoor07:book}.  These domains can
involve a large number of objects, making efficient inference a major challenge. Lifted probabilistic inference
methods address this problem by exploiting symmetries present in the
structure of the
model~\cite{Apsel2011,Braz2005IJCAI,Gogate2011,Jha2010,kersting09uai,Kisynski2009IJCAI,Milch2008,Poole2003,Sen2009b,Singla2008,Taghipour2012,GuyIJCAI11}. The
basic principle is to identify ``interchangeable''
groups of objects and perform an inference operation once per
group instead of once per individual in the group.
Researchers have proposed ``lifted'' versions of many standard propositional inference 
algorithms, including variable elimination~\cite{Braz2005IJCAI,Milch2008,Poole2003}, belief propagation~\cite{kersting09uai,Sen2009b,Singla2008}, recursive conditioning~\cite{Poole2011}, weighted model counting~\cite{Gogate2011} and knowledge compilation~\cite{GuyNips11,GuyIJCAI11}. 

Despite the progress made, we have far less insight into lifted
inference methods than into their propositional counterparts. 
Only recently has a definition been proposed for lifted inference. \emph{Domain-lifted} inference requires the time-complexity of inference
to be at most polynomial in the domain size (number of objects) of the model~\cite{GuyNips11}. In contrast, standard propositional inference is typically exponential in the domain size in probabilistic logical models.
Given this definition, it is possible to theoretically characterize, in the form of completeness results, which classes of models always permit lifted inference.
\emph{Weighted first-order model counting (WFOMC)} is the first
lifted algorithm shown to be complete for a non-trivial model
class. 
Van den Broeck~\cite{GuyNips11} showed that WFOMC is domain-lifted complete for 2-logvar
models. These models can express many important regularities, that commonly occur in
real-world problems such as (anti-)homophily and symmetry.  

We continue this line of theoretical work by analyzing \emph{lifted
variable elimination
(LVE)}~\cite{Apsel2011,deSalvoBraz2007,Milch2008,Poole2003}. We
advance the state of the art in LVE in two ways.  First, we propose a
novel LVE operator called \emph{group inversion} that generalizes the
basic inversion operator. 
This operator expands the
situations where LVE exploits the symmetry between interchangeable parts of the model, and hence increases the number of operations that can be done efficiently, via lifting. 
Second is our main result: we prove that augmenting LVE with group inversion makes it complete for the same class of models as WFOMC, namely 2-logvar models. 
This theoretical result establishes a connection between LVE and WFOMC, and also shows the importance of the the new operator.
  
The paper is structured as follows. In Section~\ref{sec:rep} and~\ref{sec:c-fove} we give the necessary background on representation and algorithms. In Section~\ref{sec:inversion} and~\ref{sec:extension} we introduce and extend the new group inversion operator. We prove completeness of LVE with the new operator in Section~\ref{sec:completeness}, and conclude in Section~\ref{sec:discussion}.

%% file: representation.tex
Probabilistic logical models combine graphical models with elements of first-order logic. Many representation languages exist for such models \cite{Getoor07:book}. Like earlier work on LVE \cite{Poole2003,deSalvoBraz2007,Milch2008}, we represent the model with \emph{parametric factors}. This formalism can compactly represent undirected probabilistic models on large numbers of objects. 
We now introduce the necessary terminology.
 
We use the term ``variable'' in both the logical and probabilistic sense. We use {\em logvar} for logical variables and {\em randvar} for random variables.  We write variables in upper- and values in lowercase. 
 
We consider factorized probabilistic models. A {\em factor} $f=\phi_f({\mathcal A}_f)$, where ${\mathcal A}_f = (A_1, \ldots, A_n)$ are randvars and $\phi_f$ is a \emph{potential} function, maps configurations of ${\mathcal A}_f$ to a real-number.  An \emph{undirected model} is a set of factors $F$ over randvars ${\mathcal A}=\bigcup_{f \in F} {\mathcal A}_f$ and represents the following probability distribution: ${\mathcal P}_F({\mathcal A}) = \frac{1}{Z} \prod_{f \in F} \phi_f({\mathcal A}_f)$, with $Z$ a normalization constant.
  
Our representation compactly defines a set of factors, using concepts of first-order logic. A {\em constant} represents an object in our universe.  A {\em term} is either a constant or a logvar.  A {\em predicate} $P$ has an arity $n$ and a finite range ($range(P)$); it maps $n$-tuples of objects (constants) to the range.  An {\em atom} is of the form $P(t_1, t_2, \ldots, t_n)$, where the $t_i$ are terms. A ground atom is an atom $P(c_1,\dots,c_n)$ where the $c_i$ are constants. The range of such a ground atom is $range(P)$. Each ground atom represents a randvar (e.g. $BloodType(joe)$). Note that the range of predicates, and hence randvars, is not limited to $\{true, false\}$ as in logic (e.g., $range(BloodType(joe)) = \{a,b,ab,o\} $).

Each logvar $X$ has a finite domain, $D(X)$, which is a set of constants $\{x_1, \dots, x_n \}$. A \emph{substitution}, $\theta=\{X_1\rightarrow t_1,\dots,X_n\rightarrow t_n\}$, is a mapping of logvars to terms. A \emph{grounding substitution} maps all logvars to constants. Applying $\theta$ to $a$, denoted $a\theta$, replaces each occurrence of $X_i$ in $a$ with $t_i$.
 
A \emph{constraint} $C_{{\mathbf X}}$ on a set of logvars ${\mathbf X}= \{X_1,\dots,X_n\}$ is a conjunction of inequalities of the form $X_i \neq t$ where $t$ is a constant in $D(X_i)$ or a logvar in ${\mathbf X}$ (the conditions $X_i \in D(X_i)$ are left implicit). We write $C$ instead of $C_{{\mathbf X}}$ when ${\mathbf X}$ is apparent from the context. By $gr({\mathbf X} | C_{{\mathbf X}})$ we denote the set of ground substitutions to ${\mathbf X}$ that are consistent with $C_{{\mathbf X}}$.

A \emph{parametrized randvar (PRV)} is a constrained atom of the form $P({\mathbf X}) |  C$, where $P({\mathbf X})$ is an atom and $C$ is a constraint on ${\mathbf X}$.
A PRV $P({\mathbf X}) |  C$ represents a set of ground atoms, and hence a set of randvars, 
$\{P({\mathbf X}) \theta | \theta \in gr({\mathbf X} | C)\}$. Given a PRV ${\mathcal V}$, $RV({\mathcal V})$ denotes the set of randvars it represents. 

\textbf{Example.}
The PRV ${\mathcal V}=Smokes(X)|X \neq x_1$, with $D(X)= \{x_1, \dots, x_n \}$, represents $n-1$ randvars $\{Smokes(x_2), \dots Smokes(x_n)\}$. \exEnd
 
A {\em parametric factor} or {\em parfactor} is of the form $\forall {\mathbf L}:C.\phi({\mathcal A})$, 
with ${\mathbf L}$ a set of logvars, $C$ a constraint on ${\mathbf L}$, ${\mathcal A} = (A_i)_{i=1}^n$ a sequence of atoms parametrized with ${\mathbf L}$, and $\phi$ a potential function on ${\mathcal A}$. The set of logvars occurring in ${\mathcal A}$ is denoted $logvar({\mathcal A})$, and we have $logvar({\mathcal A})\subseteq{\mathbf L}$. When $logvar({\mathcal A})={\mathbf L}$, we write the parfactor as $\phi({\mathcal A})|C$. A factor $\phi({\mathcal A}')$ is called a {\em grounding} of a parfactor $\phi({\mathcal A})|C$ if ${\mathcal A}'$ can be obtained by instantiating ${\mathbf L}$ according to a grounding substitution $\theta \in gr({\mathbf L}|C)$. The set of all groundings of a parfactor $g$ is denoted $gr(g)$.

\textbf{Example.}
Parfactor $g= \phi(Smokes(X), Asthma(X))$ 
represents the set of $n$ ground factors $gr(g)=\{\phi(Smokes(x_1),Asthma(x_1)), \dots, \phi(Smokes(x_n),Asthma(x_n)) \}$. \exEnd

When talking about a \emph{model} below, we mean a set of parfactors. In essence, a set of parfactors $G$ is a compact way of defining a set of factors $F = \{ f | f \in gr(g) \land g \in G\}$. The corresponding probability distribution is ${\mathcal P}_G({\mathcal A}) = \frac{1}{Z} \prod_{f \in F} \phi_f({\mathcal A}_f)$.

%% file: c-fove.tex
The state of art in lifted variable elimination (LVE) is the result of various complementary efforts~\cite{Poole2003,Milch2008,Apsel2011,deSalvoBraz2007}. This section reviews the most recent algorithm for LVE, namely \mbox{C-FOVE}~\cite{Milch2008}. 

Variable elimination calculates the marginal distribution of some variable by \emph{eliminating} randvars in a specific order from the model until reaching the desired marginal~\cite{PooleZ03}. To eliminate a single randvar $V$, it first \emph{multiplies} all factors containing $V$ into a single factor and then \emph{sums out} $V$ from that single factor. LVE does this on a lifted level by eliminating parametrized randvars (i.e., whole groups of randvars) from parfactors (i.e., groups of factors). The outer loop of LVE is as follows. 
 
\begin{center}
\begin{tabular}{l}
\hline
\textbf{Inputs}: $G$: a model; $Q$: the query randvar.\\
\emph{while} $G$ contains other randvars than $Q$:\\
\quad \emph{if} a PRV $\mathcal{V}$ can be eliminated by lifted sum-out\\
\qquad $G \leftarrow$ eliminate $\mathcal{V}$ in $G$ by lifted sum-out\\
\quad \emph{else} apply an enabling operator on parfactors in $G$\\
\emph{end while}\\
\emph{return} $G$\\
\hline
\end{tabular}
\end{center}
 
As this shows, LVE works by applying a set of lifted {\em operators}. We now discuss the most basic operators. Beside these, LVE has \emph{conversion} operators, which we discuss in Section~\ref{sec:extension}.  

\textbf{Lifted Sum-out}. This operator sums-out a PRV, and hence all the randvars represented by that PRV, from the model. Lifted sum-out is applicable only under a precondition (each randvar represented by the PRV appears in exactly one grounding of exactly one parfactor in the model).  
The goal of all other operators is to manipulate the parfactors into a form that satisfies this precondition.  In this sense, all operators except lifted sum-out can be seen as \emph{enabling operators}.

\textbf{Lifted Multiplication}. This operator performs the equivalent of many factor multiplications in a single lifted operation. It prepares the model for sum-out by replacing all the parfactors that share a particular PRV by a single equivalent product parfactor in the model.

\textbf{Splitting and Shattering}.
These operators rewrite the model such that a pair of atoms or formulas represent either identical or disjoint groups of randvars.

%% file: inversion.tex
We now introduce a new lifted operator called \emph{group inversion}.  This operator generalizes the existing inversion operator of FOVE~\cite{Braz2005IJCAI,Poole2003} and is inspired by the concept of disconnected groundings in lifted recursive conditioning~\cite{Poole2011}. We first review the existing inversion operator, and then define group inversion. The motivation behind adding this new operator to LVE is that it makes LVE complete for important classes of models, see Section~\ref{sec:completeness}.

\subsection{Inversion}

Lifted sum-out eliminates a PRV, i.e., a whole group of randvars, 
in a single operation.  An important principle that it relies on is \emph{inversion} \cite{Poole2003,deSalvoBraz2007}.

Inversion consists of turning a sum of products into a product of sums. 
Consider the sum of products $\sum_i \sum_j i \cdot j$. 
If the range of $j$ does not depend on  $i$, it can be rewritten as $(\sum_j j) (\sum_i i)$, which is a product of sums. 
More generally, given $n$ variables $x_1, \dots, x_n$, with independent ranges, we have
\[ \sum_{x_1} \sum_{x_2} \ldots \sum_{x_n} \prod_i f(x_i) = \prod_i \sum_{x_i} f(x_i). \]
Furthermore, if all $x_i$ have the same the range, this equals $(\sum_{x_1} f(x_1))^n$. That is, the summation can be performed for \emph{only one} representative $x_1$ and the result used for all $x_i$.  

Exactly the same principle can be applied in lifted inference for summing out randvars.  Suppose we need to sum out  
$RV(F(X_1,X_2))$ from the parfactor $g= \phi(F(X_1,X_2), P(X_1,X_2))$.  For each instantiation $(x_1, x_2)$ of $(X_1, X_2)$, $F(x_1, x_2)$ has the same range, hence applying inversion yields
\begin{eqnarray}
\label{eq:inversion}
\sum_{F(x_1,x_1)}\sum_{F(x_1,x_2)} \dots \sum_{F(x_n,x_n)} \prod_{\theta \in \Theta}  g \theta 
=  \prod_{\theta \in \Theta} \big(\sum_{F(X_1, X_2)\theta} g\theta \big)
\end{eqnarray}
with $\Theta=gr(X_1,X_2)$. This shows that we can perform the sum-out operations independently for each $F(X_1,X_2)\theta$, and multiply the results. Furthermore, since all the factors $g\theta$ are groundings of the same parfactor and have the same potential $\phi$, the result of summing out their first argument $F(X_1,X_2) \theta$ is also the same potential, denoted $\phi'$. It thus suffices to only perform one instance of these sum-out operations and rewrite Expression~\ref{eq:inversion} as 
\begin{eqnarray*}
&&\prod_{\theta \in \Theta} \Big(\sum_{F(X_1, X_2)\theta} \phi(F(X_1,X_2)\theta,P(X_1,X_2)\theta)\Big)
= \prod_{\theta \in \Theta} \Big(\phi'(P(X_1,X_2)\theta)\Big) = gr(g') \nonumber
\end{eqnarray*} 
where $g' = \phi'(P(X_1,X_2))$. This is what lifted sum-out by inversion does: it directly computes parfactor $g'$ from $g=\phi(F(X_1,X_2),P(X_1,X_2))$ by summing out $F(X_1,X_2)$ from $g$ in a single operation. This single lifted operation replaces $|\Theta|$ sum-out operations on the ground level.

\subsection{Group Inversion: Principle}
Inversion only works when 
the summations are independent. Our first contribution is based on the following observation. 
When we cannot apply inversion because of dependencies between factors, we can still partition the factors (and the summations) into groups such that dependencies exist only among factors within a group, but not between groups. Furthermore, we can compute the result for one group and use it for all groups, provided that these groups are {\em isomorphic}, i.e., that there exists a one-to-one-mapping of the randvars from one group to the others such that exactly the same sum of products is obtained. We call this  \emph{group inversion}.

Consider the parfactor $g= \phi(F(X_1,X_2), F(X_2,X_1)) | X_1 \neq X_2$ and assume we want to sum out the randvars $RV(F(X_1,X_2) | X_1 \neq X_2)$. If we focus on the part of the computation related to one particular instantiation $(x_1,x_2)$, the sum looks as follows.
\begin{eqnarray*}
\ldots \sum_{F(x_1, x_2)} \ldots \sum_{F(x_2,x_1)} \ldots
\Big( \phi(F(x_1,x_2), F(x_2,x_1)) \cdot \phi(F(x_2,x_1), F(x_1,x_2) \Big) \cdot (\ldots)
\end{eqnarray*}
The product contains two factors over the considered pair of randvars $F(x_1,x_2)$ and $F(x_2,x_1)$. The product of these two cannot be moved out of the summation over either of the two randvars. Still, the two summations are independent of all other factors, 
and can be isolated from the rest of the computation. The same can be done for each pair of instantiations $(x_i, x_j)$ of $(X_1,X_2)$. 
This means that summing out $RV(F(X_1,X_2) | X_1 \neq X_2)$ from $g$ can be done using group inversion:

\begin{eqnarray*}
\sum_{F(x_1,x_2)}\sum_{F(x_1,x_3)} \dots \sum_{F(x_{n-1},x_n)} \Big( \prod_{\theta_{ij} \in \Theta}  g \theta_{ij} \Big) 
= \prod_{\{\theta_{ij}, \theta_{ji}\} \in {\Theta}} \Big(\sum_{F(X_1,X_2)\theta_{ij}} \sum_{F(X_1,X_2)\theta_{ji}} g \theta_{ij} \cdot g \theta_{ji} \Big)
\end{eqnarray*}
where $\theta_{ij}$ is a grounding substitution $\{X_1 \rightarrow x_i,  X_2 \rightarrow x_j\}$ in $\Theta=gr(X_1,X_2|X_1\neq X_2)$. Lifting is now possible again because for all distinct pairs of substitutions $(\theta_{ij}, \theta_{ji})$  in $\Theta$, the pairs of factors $(g \theta_{ij}, g \theta_{ji})$ share the same potential $\phi$. As such, the multiplicands of each pair also have the same potential $\phi'$, and summing out their arguments results in the same potential $\phi''$. Hence, it suffices to perform only one (lifted) instance of these operations as follows
\begin{eqnarray*}
&&\prod_{\{\theta_{ij}, \theta_{ji}\} \in {\Theta}} \Big( \sum_{F(X_1,X_2)\theta_{ij}} \sum_{F(X_1,X_2)\theta_{ji}} \phi'(F(X_1,X_2 )\theta_{ij}, F(X_1,X_2 ) \theta_{ji}) \Big)\\ 
&=& \prod_{\{\theta_{ij}, \theta_{ji}\} \in  \Theta} \phi''() = \prod_{\theta_{ij} \in  \Theta} \phi''()^{1/2} = gr(g'),
\end{eqnarray*}
where $g'$ is the parfactor $\forall X_1,X_2:X_1 \neq X_2. \phi''()$ with $\phi''$ a potential function with no arguments (i.e., a constant, because both arguments have been summed-out). This is what group inversion does.

\begin{figure}
\begin{center}
\includegraphics[width = 0.85\linewidth ]{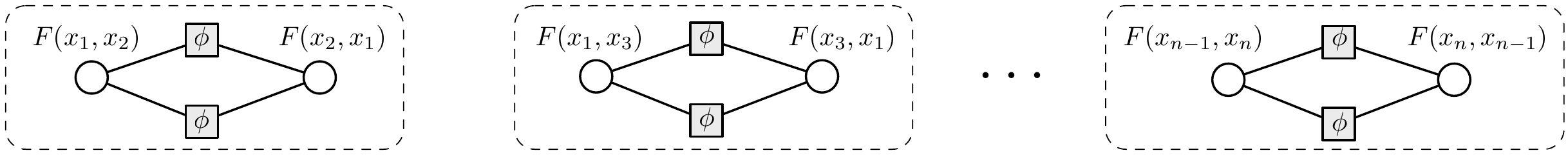}
\end{center}
\caption{Group inversion on pairs of randvars. Circles represent randvars, squares represent factors. Dashed boxes indicate the partitioning into groups.} \label{fig:group-inv}
\end{figure} 

Group inversion partitions the set of factors (and randvars) into independent and isomorphic groups. An important question is what such a partitioning looks like. Figure~\ref{fig:group-inv} shows this for the above example. 
In general, let us call two factors {\em directly linked} if they share a randvar, and let {\em linked} be the transitive closure of this relation. Factors that are linked end up in the same group. 
Sometimes this yields useful partitionings, sometimes not. As a `negative' example, consider a parfactor $\phi(P(X_1), P(X_2))$. Any two ground factors $\phi(P(x_i), P(x_j))$ and $\phi(P(x_k), P(x_l))$ are linked (since both are directly linked to $\phi(P(x_j), P(x_k))$). Hence the only option is the trivial partition in which all ground factors are in a single, large group, which is not practically useful. As a `positive' example, consider the case where each atom uses all the logvars in the parfactor, as in the earlier example $\phi(F(X_1,X_2), F(X_2, X_1))| X_1 \neq X_2$.  In such cases, we can always partition the randvars into groups such that the size of a group is independent of the domain size. The reason is that in such cases, the arguments of the atoms in a linked group are necessarily \emph{permutations} of each other. Hence the size of a linked group can be no larger than the number of possible permutations, which is independent of the domain size. We use this property in our group inversion operator. 

\subsection{The Group Inversion Operator}
\label{sec:inv-op}

\textbf{In- and output.} Group inversion takes a parfactor $g=\phi(\mathcal{A})|C$ and a set of atoms $\{A_1,\dots,A_n\} \subseteq \mathcal{A}$ as input. It returns a new parfactor that is the result of summing out $\{A_1, \dots, A_n\}$ from $g$.

\textbf{Preconditions.} Group inversion is applicable when 
\textbf{(i)} for all $i,j$: $RV(A_i|C) = RV(A_j|C)$, \textbf{(ii)} each $A_i$ has all the logvars ${\mathbf L}$ in the parfactor, \textbf{(iii)} for each pair of logvars $X_i,X_j\in {\mathbf L}$ there is an inequality constraint $X_i \neq X_j$ in $C$, and \textbf{(iv)} for each PRV ${\mathcal V}$ outside $g$: $RV(A_i|C)\cap RV({\mathcal V}) =\emptyset$.  The key observation is that, in such a parfactor, due to conditions (i) and (ii), for each $i\neq j$, randvar $A_i$ is a permutation of  $A_j$. That is, $\lambda_{ij}(A_i) = A_j$, where $\lambda_{ij}$ is a permutation of the logvars and $\lambda(A_i)$ represents the result of applying $\lambda$ on the arguments of $A_i$. 

\textbf{Operator.} When the preconditions hold, group inversion applies the following four steps. We further explain these steps below. 

\begin{enumerate}
\item \emph{Partition.} Find the set $\Lambda$ of permutations $\lambda_{ij}$ such that $\lambda_{ij}(A_i)=A_j$. 
Then find the closure $[\Lambda]$ of $\Lambda$, i.e., the minimal set $[\Lambda]$ of permutations such that $\Lambda \subseteq [\Lambda]$, and $[\Lambda]$ is closed under composition.
\item \emph{Multiply} to compute the parfactor $g_{[\Lambda]} = \phi'({\mathcal A}')|C$ as the product $\prod_{\lambda \in [\Lambda]} g_{\lambda}$, where $g_{\lambda} = \phi(\lambda({\mathcal A}))|C$.
\item \emph{Sum-out} to compute $g'$ $=$ $\phi''({\mathcal A}'')|C$ $=$ $\sum_{{\mathcal A}_{[\Lambda]}} g_{[\Lambda]}$, where the summed out atoms ${\mathcal A}_{[\Lambda]}$ $=$ $\{A'_1 \dots, A'_m\}$ are such that $\forall i,j: RV(A_i|C)=RV(A'_j|C)$.
\item \emph{Scale.} Return $g'' = \phi''({\mathcal A}'')^{1/ m}|C$, with $m=|[\Lambda]|$ (or equivalently: $m=|{\mathcal A}_{[\Lambda]}|$).
\end{enumerate}

\textbf{Step1 (partition).} The goal here is to find, on the lifted level, the set of factors and randvars that need to be put (and summed-out) in the \emph{same group}. 
In $gr(g)$, each pair of factors $(g\theta,g\theta')$ that are directly linked 
can be derived from each other by a \emph{permutation} of constants. Concretely, if randvar $A_i \theta$ in $g\theta$ is the same as $A_j \theta'$ in $g\theta'$, then we have $\theta = \lambda_{ij}(\theta')$.\footnote{Note that for any atom $A_i$, and a permutation $\lambda$ of its logvars, $A_i \lambda(\theta) = \lambda(A_i) \theta$.} 
All factors that are directly linked to a factor $g\theta$ are thus in the set $\{ g \theta' | \theta' = \lambda(\theta) , \lambda \in \Lambda \}$. In Step 1, we find the set of permutations $\Lambda$ that convert directly linked factors to each other. 
Since directly linked factors are derived from each other by a permutation in $\Lambda$, each factor linked to $g\theta$ can be written as $g \lambda(\theta)$, where $\lambda$ is a \emph{composition} of permutations in $\Lambda$. We can thus find all such factors by computing the \emph{closure} of $\Lambda$ under the operation of composition, this is also called the minimal permutation group. That is, the closure is the minimal set $[\Lambda] \supseteq \Lambda$, such that $\forall \lambda_1, \lambda_2 \in [\Lambda]: \lambda_1.\lambda_2 \in [\Lambda]$. 
Note that all the permutations are computed on the lifted level, i.e., as permutations of the logvars.
 
\textbf{Example.} Consider finding the closure $[\Lambda]$ for a parfactor with atoms $A_1=F(W,X,Y,Z)$ and $A_2=F(Z,W,X,Y)$. Note that a right-shift of $A_1$'s logvars yields $A_2$. Let us denote a permutation $\lambda = \{W\rightarrow W',\dots Z\rightarrow Z'\}$ as $\lambda=(W',\dots,Z')$. For atoms $\{A_1,A_2\}$, $\Lambda$ includes the three 
permutations $\lambda_{11}=\lambda_{22}=(W,X,Y,Z)$, $\lambda_{12}=(Z,W,X,Y)$ and $\lambda_{21}=(X,Y,Z,W)$ (respectively identity, right- and left-shift). $\Lambda$ is not closed under composition because, for example, the composition $\lambda'=\lambda_{12}.\lambda_{12}=(Y,Z,W,X)$ is not in $\Lambda$. 
Adding $\lambda'$ to $\Lambda$, however, yields a closed set of permutations, as no other possible composition results in a permutation out of this set. As such, we have $[\Lambda]=\Lambda \cup \{\lambda'\}$. Note that $[\Lambda]$ includes only 4 of all the possible $4!=24$ permutations. \exEnd

\textbf{Step 2 (multiply).} Next we apply lifted multiplication to compute a parfactor $g_{[\Lambda]} = \prod_{\lambda \in [\Lambda]} g_{\lambda}$, where $g_{\lambda} = \phi(\lambda({\mathcal A}))|C$. This parfactor has the same form as the product of all the factors in a representative group (for a given order of multiplications). 

\textbf{Step 3 (sum-out).} Next we perform lifted sum-out, which sums-out the set of atoms $\{\lambda(A_i) | \lambda \in [\Lambda]\}$ from parfactor $g_{[\Lambda]}$ and yields parfactor $g'$. At this point, all randvars $RV(A_i|C)$ have been eliminated from the model, which is the goal of the entire group inversion procedure. 

\textbf{Step 4 (scale).} For correctness, we still need to scale $g'$, as it represents multiple equivalent factors. Since $g'$ has the same constraint as $g$ and hence the same set of all possible grounding substitutions $\Theta$, $gr(g')$ represents $|[\Lambda]|$ 
equivalent factors instead of one factor, for each group of the factors in $gr(g)$.\footnote{This equivalence follows from the fact that each of them can be derived from a same set of factors, but with a different order of multiplications in Step 2.} Hence, we replace the potential $\phi''$ of $g'$ with $\phi^{''1/|[\Lambda]|}$ to preserve the distribution.

\textbf{Remarks.} 
Since group inversion includes parfactor multiplication, beside removing the summed-out atoms, it can add to the resulting parfactor additional arguments that result from permutation of logvars in its original arguments. The following example depicts this.

\textbf{Example.} Consider applying group inversion to eliminate the set of randvars $RV(F(X,Y))$ from $g= \phi(S(X), F(X,Y), F(Y,X), A(Y))|X \neq Y$. In Step 1, we find the permutation group $[\Lambda]$ for $A_1=F(X,Y)$ and $A_2=F(Y,X)$. The group ${[\Lambda]} = \{\lambda,\lambda'\}$ consists of the identity permutation $\lambda = \{X \rightarrow X, Y \rightarrow Y \}$, and the permutation $\lambda' = \{X \rightarrow Y, Y \rightarrow X \}$. In Step 2, we multiply $g_{\lambda} = g$ and $g_{\lambda'} = \phi(S(Y), F(Y,X), F(X,Y), A(X))|X \neq Y$, to compute $g_{[\Lambda]} = \phi'(S(X),A(X),F(X,Y),F(Y,X),S(Y),A(Y))|X \neq Y$. In Steps 3 and 4, we respectively sum-out  atoms $F(X,Y), F(Y,X)$ from $g_{[\Lambda]}$ and scale the resulting potential, to compute the parfactor $g'' = \phi''(S(X),A(X),S(Y),A(Y))^{1/2}|X \neq Y$. Note that $g''$ includes atoms $A(X)$ and $S(Y)$, which were not present in the original parfactor $g$. 
\exEnd

As mentioned before, group inversion has inversion as a special case (namely when there is only one atom 
in the parfactor that covers the randvars $RV(A_i|C)$, i.e., when $n=1$ in the operator). In this case, the closure $[\Lambda]$ consists only of the identity permutation. 
 
\begin{theorem}
\label{theo:ginv}
Lifted sum-out with the group inversion operator is equivalent to summing out the randvars on the ground level.
\end{theorem}
The proof is provided in the appendix (it relies on showing that the corresponding ground operations are independent and isomorphic).

%% file: extension.tex
LVE exploits symmetry among groups of interchangeable and \emph{independent} objects, as we have seen in group inversion. Additionally, LVE also exploits symmetry within groups of interchangeable but \emph{dependent} objects, by \emph{counting}. Inspired by counting elimination~\cite{deSalvoBraz2007}, Milch et al.\ introduced \emph{counting formulas} and operators that handle them in the C-FOVE algorithm~\cite{Milch2008}. This was later extended by Apsel and Brafman~\cite{Apsel2011}. These operators play a key role in LVE and in deriving our completeness results (Section~\ref{sec:completeness}) . Below we first review counting formulas and the existing operators that handle them. Then we extend group inversion with support for summing-out counting formulas.

\textbf{Counting formulas.} A counting formula is of the form $\gamma = \#_{X:C}[P({\mathbf X})]$, with $C$ a constraint on the \emph{counted logvar} $X \in {\mathbf X}$ (if the only condition is $X \in D(X)$, we do not write out $C$). Because the counted logvar $X$ is already bound by the counting formula, it is excluded from $logvar(\gamma)$. A grounded counting formula is a counting formula in which all logvars except the counted logvar are replaced by constants. Such a formula represents a \emph{counting randvar} (CRV). The range of the CRV is the set of possible histograms of the form $h= \{(r_i, n_i)\}_{i=1}^{|range(P)|}$, that shows for each $r_i \in range(P)$ the number $n_i$ of randvars $P(\dots, x, \dots)$ whose state is $r_i$. The state of the counting randvar thus depends deterministically on the state of the randvars $RV(P(\dots, X, \dots) | C)$. 

\textbf{Example.}
$\#_{Y}[F(X,Y)]$ is a counting formula and $\#_{Y}[F(x_1,Y)]$ a grounded counting formula that counts the number of friends of $x_1$. Having $D(Y)= \{y_1, y_2, y_3\}$, the grounded counting formula covers the randvars $\{F(x_1,y_1), F(x_1,y_2), F(x_1,y_3)\}$ and defines a CRV, the value of which is determined by the values of these three randvars. For instance, if $F(x_1,y_1) = true$, $F(x_1,y_2) = false$ and $F(x_1,y_3)=true$, the value of the CRV is the histogram $\{(true,2),(false,1)\}$.

Counting formulas can be introduced in the model by the following two conversion operators.

\textbf{(Just-different) Counting Conversion} \cite{Milch2008,Apsel2011}. This operator introduces counting formulas in the model, to exploit the symmetry between a group of \emph{interchangeable} randvars. By replacing an atom such as $A(X)$ with a counting formula $\#_X[A(X)]$, we compactly represent and manipulate a single potential on randvars $A(x_1), \dots, A(x_n)$. Intuitively, this conversion achieves the equivalent of multiplying groundings of a \emph{single} parfactor with each other. This operator is applicable on a set of logvars that only appear in a single atom or in \emph{just-different} atoms, that is, pairs of atoms $P(X_1,{\mathbf X}),P(X_2,{\mathbf X})$ whose logvars are constrained as $X_1 \neq X_2 $, such as atoms $A(X_1)$ and $A(X_2)$ in the parfactor $\phi(A(X_1),A(X_2))|X_1\neq X_2$. 
 
\textbf{Joint Conversion} \cite{Apsel2011}. This 
operator is an enabling operator for counting conversion. Joint conversion on a pair of atoms $A(X),B(X)$ replaces any occurence of $A(.)$ or $B(.)$ with a \emph{joint} atom $J_{AB}(.)$, whose range is the Cartesian product of the range of $A$ and $B$. Such a conversion enables counting conversion, when it results in a model with just-different joint atoms.

\textbf{Example.} Consider the parfactor $\phi(S(X),A(X),S(Y),A(Y)) |X \neq Y$. By joint conversion on atoms $S(.)$ and $A(.)$, we rewrite this parfactor as $\phi'(J_{S A}(X),J_{S A}(X),J_{S A}(Y),J_{S A}(Y)) |X \neq Y$, which is trivially simplified to $\phi''(J_{S A}(X), J_{S A}(Y)) |X \neq Y$. Note that $J_{S A}(X)$ and $J_{S A}(Y)$ are just-different atoms. By just-different counting conversion, we rewrite this parfactor into the form $\phi^{\#}(\#_X[J_{S A}(X)])$. This parfactor is now prepared for application of lifted sum-out. 

\textbf{Extension of Group Inversion.} The above are existing operators. Now that we have counting formulas, we also need to support them in our new group inversion operator. Counting formulas, like atoms, can be eliminated by lifted sum-out. This can be done by group inversion with a small modification of the operator. When $\{A_1, \dots, A_n\}$ is a group of counting formulas of the form $A_i=\#_{X_i:C'}[P(X_i;{\mathbf{L}})]$, the operator eliminates all the formulas by following the same 4 steps as in Section~\ref{sec:inv-op}, with the exception of the \emph{sum-out} step (Step 3), which becomes: 
$$\sum_{(h_1, \dots, h_m) \in range(A'_1, \dots, A'_m)} \Big( \big(\prod_{i=1}^m \textsc{num}(h_i)\big) \; g_{[\Lambda]} \Big)$$
where, for a histogram $h_i=\{(r_i,n_i)\}_{i=1}^{r}$ with $\sum_i n_i =n$, the coefficient $\textsc{num}(h_i)$ is the multinomial coefficient $\frac{n!}{\prod_i n_i!}$ representing the number of possible assignments to $RV(A'_i|C)$ that yield this histogram. This operator generalizes the existing sum-out operation of LVE~\cite{Milch2008}. 

%% file: completeness.tex
We now show that including our group-inversion operator yields certain completeness results. These are the first such results for LVE and the second for exact lifted methods in general (after WFOMC \cite{GuyNips11}). The concrete LVE algorithm considered is \mbox{C-FOVE} \cite{Milch2008} extended with group-inversion and joint formulas~\cite{Apsel2011}. We refer to it as C-FOVE$^{+}$. We use the definitions of Van den Broeck \cite{GuyNips11}.
\begin{definition} [Domain-lifted algorithm]
A probabilistic inference algorithm is \emph{domain-lifted} for a model $G$, query ${\mathcal Q}$ and evidence ${\mathcal E}$ iff it runs in polynomial time in $|n_1|, ..., |n_k|$, with $n_i$ the domain of logvar $X_i \in logvar(G,{\mathcal Q},{\mathcal E})$.
\end{definition}
\begin{definition} [Completeness]
An algorithm is \emph{complete} for a class ${\mathcal M}$ of models if the algorithm is domain-lifted for all models $G \in {\mathcal M}$ and all ground queries ${\mathcal Q}$ and evidence ${\mathcal E}$.
\end{definition}

\subsection{Main Result: Completeness for 2-logvar models}

We show that C-FOVE$^{+}$ is a complete domain-lifted algorithm for the subclass of parfactor models called $2$-logvar models. A parfactor model is \emph{2-logvar} if no parfactor has more than 2 different logvars, and the model contains no counting formulas. We discuss the importance of this class of models in the next section.

\begin{theorem}
C-FOVE$^{+}$ is a complete domain-lifted algorithm for 2-logvar models.
\end{theorem}
\emph{Proof.} We show the proof here since it is `constructive': it shows how C-FOVE$^+$ deals with 2-logvar models: first sum-out all 2-logvar atoms (atoms $A$ with $|logvar(A)|$ $=$ $2$), then 1-logvar atoms, then 0-logvar atoms. We then show that this procedure is domain-lifted.\footnote{We assume that the model $M$ is preemptively shattered and in normalized form \cite{Milch2008,Poole2011}. Any 2-logvar model $M$ can be rewritten in poly time as an equivalent 2-logvar model $M'$ that satisfies these conditions.} 

\emph{Step 1}. We eliminate all \textbf{2-logvar} atoms in two steps. (a) We multiply the parfactors until there are no distinct pairs $(A_i,A_j)$ of 2-logvar atoms in distinct parfactors $(g_i,g_j)$, such that $RV(A_i|C_i)=RV(A_j|C_j)$. The resulting equivalent model $M^*$ is a 2-logvar model, since multiplication preserves the number of logvars in the product \cite{Milch2008}. 
(b) We eliminate each 2-logvar atom in $M^*$ by group inversion. This is possible since all $2$-logvar atoms that represent the same randvars are in the same parfactor, and have both of the logvars. The result is a 1-logvar model.

\emph{Step 2}. We eliminate all \textbf{1-logvar} atoms in four steps. (a) We (repeatedly) perform joint conversion on a pair of atoms $P_1(X_1,{\mathbf c}_1)$ and $P_2(X_2,{\mathbf c}_2)$ (of distinct predicates). 
This replaces them with joint atoms $J_{12}(X_1,{\mathbf c}_{12})$ and $J_{12}(X_2,{\mathbf c}_{12})$, respectively. 
When no more such conversions are possible, any pair of logvars $X_1,X_2$ that are constrained as $X_1\neq X_2$, appear only in pairs of just-different atoms (otherwise, a joint conversion is still possible between them). 
(b) We perform (just-different) counting conversion on all the logvars. (c) We multiply all the parfactors into one, which is trivially possible since the model contains no free logvars. In the resulting parfactor each argument is either a ground atom, or a counting formula of the form $\gamma_i=\#_{X_i}[J_{1 \dots k_i}(X_i)]$. 
(d) We sum-out the counting formulas. 
The result of this step is a model in which all arguments are ground, i.e., a 0-logvar model.

\emph{Step 3}. We eliminate all the remaining (so \textbf{0-logvar}) non-query randvars. Inference is now performed at the ground level, i.e., with standard VE. This step concludes the inference process. 

\emph{Complexity.} All the operations in Steps 1 and 2 run in time polynomial in the domain of the logvars. The most expensive step is handling the counting formulas produced in Step 2b. The largest size for the range of these formulas is $O(n^r)$ where $r$ is the largest range size among the (joint) atoms. As such the exponent $r$ is independent of the domain size. The complexity of Step 2 is thus polynomial in the domain size. Step 3 has worst case complexity $O(m^{c})$, with $c$ the total number of symbols appearing in $(M, {\mathcal Q}, {\mathcal E})$, and $m$ the largest size of range among the randvars. This complexity satisfies the definition of a domain-lifted algorithm~\cite{GuyNips11}. As such, C-FOVE$^+$ is a domain-lifted algorithm for any 2-logvar model, and hence \emph{complete} for this class of models. \exEnd
	
\subsection{Importance of the Result}
\label{sec:importance}

Our completeness result furthers our understanding of the relation between LVE and lifted search based methods, which is an important problem in the field~\cite{Gogate2011,GuyIJCAI11,Poole2011}.
Van den Broeck \cite{GuyNips11} showed that his WFOMC algorithm is complete for the class of  2-WFOMC problems. Any such problem can be represented as a $2$-logvar model, and vice versa (see the appendix). Our completeness result for LVE is thus equally strong as that of Van den Broeck for WFOMC.

The class of 2-logvar models includes many useful and often employed models in statistical relational learning. It can model multiple kinds of relations, including: \emph{homophily} between linked entities, e.g., $\phi(Property(X),Related(X,Y),Property(Y))$; 
\emph{symmetry}, e.g., $\phi(Friend(X,Y),Friend(Y,X))$; \emph{anti-symmetry}, e.g., $\phi(Smaller(X,Y),Smaller(Y,X))$; and \emph{reflexivity}, e.g., $\phi(Knows(X,X))$. 
The completeness result shows that for these models, LVE can perform inference in time polynomial in the domain size. 
An example of models that fall outside of the 2-logvar class are models containing a transitive relation, e.g.\ $\phi(Like(X,Y),Like(Y,Z),Like(X,Z))$. For such models, no domain-lifted inference procedure is known. An important direction for future work is the derivation of (positive or negative) completeness results for such model classes.   

Next to our main result (Theorem~2), we now present a second result (Theorem~3) that is in line with a known result for lifted recursive conditioning \cite{Poole2011}. This result applies to models that restrict the number of logvars per \emph{atom}, while our main result restricts the number of logvars per \emph{parfactor}.
 
\begin{theorem} 
C-FOVE$^{+}$ is a complete domain-lifted algorithm for the class of models in which each atom has at most 1 logvar. 
\end{theorem}
  
The proof is provided in the appendix (it builds on the proof of Theorem 2).

%% file: conclusion.tex
We showed how introducing a new inference operator, called group inversion, makes lifted variable elimination a complete domain-lifted algorithm for 2-logvar models.  A corollary of the completeness result is that lifted variable elimination and WFOMC are domain-lifted complete for exactly the same subclass of models. 
We believe that future research on the relationships between the various lifted inference algorithms will yield valuable theoretical insights, similar to those about the propositional inference methods~\cite{Darwiche01, Dechter07,Dechter99}.

%% file: suppl.tex
In this appendix we provide proofs for Theorem 1 and 3, and present a procedure for transforming weighted model counting (WMC) models to parfactor models.
\section{Proof of Theorem~1}


We prove the theorem by showing that the corresponding ground operations are both \emph{independent} and \emph{isomorphic}.

\textbf{Independence.} We require the following definition. 
 
Given a set of factors $F$ and set of randvars $R$, we call a subset of factors $F' \subseteq F$ \emph{mutually closed} with respect to a group of randvars $R' \subseteq R$, if (i) no factor in $F\setminus F'$ contains a randvar $r' \in R'$, (ii) no randvar in $R \setminus R'$ appears in a factor $f' \in F'$, and (iii) each randvar $r' \in R'$ appears in some factor $f' \in F'$. 

Now, we show that we can form mutually closed sets of randvars and factors in $R= RV(A_i |C)$ and $F= gr(g)$ by partitioning them into sets in which all elements are permutations of each other (can be derived from one another by a permutation of constants). The set of permutations that defines the partitioning is the minimal permutation group $[\Lambda]$. 

Given a set of permutations $\Lambda$ on ${\mathbf X}$, we define two substitutions $\theta_1,\theta_2$ to be in the relation $\sim_\Lambda$ iff $\lambda(\theta_1) = \theta_2$ for some $\lambda \in \Lambda$. Using this relation we can define a partitioning of a set of substitutions $\Theta$ as ${\mathbf \Theta}_\Lambda$, where $\theta$ and $\theta'$ are in the same group if and only if $\theta \sim_\Lambda \theta'$. 

As shown in steps 1 and 2 of the operator, for any two factors $g \theta$ and $g\theta'$ that share a randvar from the set $RV(A_i)$, we have $\theta = \lambda(\theta')$, for some $\lambda \in [\Lambda]$. Thus for any $\Theta_i \in {\mathbf \Theta}_{[\Lambda]}$, the set of factors $F_i = \{g \theta | \theta \in \Theta_i\}$ are mutually closed w.r.t.\ the set of randvars $R_i = \{ A_i \theta | \theta \in \Theta_i\}$. This shows that we can divide the problem of summing out $RV(A_i)$ from $gr(g)$ into \emph{independent} problems of summing out each set of randvars $R_i$ from the set of factors $F_i$.

\textbf{Isomorphism.} We show that the sum-out problems are also \emph{isomorphic}, by a mapping between the ground substitutions that produce ground factors in each group.

To show the isomorphism between groups of $gr(g)$, we note that each group is formed from the factors $\{ g \theta | \theta \in \Theta_i \}$, where $\Theta_i$ is a group in ${\mathbf \Theta}_{[\Lambda]}$. The one-to-one mapping between the factors can thus be established by a one-to-one mapping between the constants of the grounding substitutions in different groups $\Theta_i$ and $\Theta_j$. This is done by starting from an arbitrary pair of substitutions $\theta_i \in \Theta_i$ and $\theta_j \in \Theta_j$ and mapping the constants that are assigned to the same logvar to each other. It follows then that each substitution $\theta'_i \in \Theta_i$ such that $\lambda(\theta_i) = \theta'_i$ is mapped to exactly one substitution $\theta_j' \in \Theta_j$ such that $\lambda(\theta_j) = \theta'_j$. As such the set of factors (and the set of randvars) are isomorphic up to a renaming of 
the constants in each group.

This shows that the sum-out problems in different groups are independent and isomorphic. Hence, it is correct to replace them by a single lifted operation, i.e.\ to solve one instance of the problem for a representative group and generalize the result for all, as is performed in lifted sum-out by the group inversion operator.  

\section{Proof of Theorem~3}


\emph{Proof sketch.} The proof builds on the proof of Theorem 2 (given in the paper). Note that the approach used in Steps 2 and 3 of the proof of Theorem 2 is also applicable here. The operations in Step 2, which together eliminate the $1$-logvar atoms, do not depend on the total number of logvars in the parfactors. Using this approach, we can eliminate all the $1$-logvar atoms in any model whose atoms contain at most one logvar. The resulting model can be solved as in Step 3. As was shown in the proof of Theorem 2, the time-complexity of these steps is polynomial in the domain size. The inference procedure is thus domain-lifted. \exEnd

\section{Transformation from WMC to Parfactor Models}
In this section we introduce a method for transforming any weighted model counting (WMC) model~\cite{GuyIJCAI11,GuyNips11,Gogate2011} to an equivalent parfactor model~\cite{Milch2008,Poole2003,deSalvoBraz2007}, i.e., a transformation from the representation used by WFOMC to the representation used by LVE. 

A WMC model $M = (\mathcal{C}, w)$ consists of a set of constrained clauses $\mathcal{C}$ and a weight function $w$ that maps each predicate $P$ to a \emph{weight} $w(P)$. We present a transformation from such a model to an equivalent parfactor model. Given any $k$-WFOMC model (with clauses containing up to $k$ logvars), the following transformation method returns an equivalent $k$-logvar parfactor model (with parfactors containing up to $k$ logvars).

Consider a WMC model $M$ with the weighting function $w$ and the set of constrained clauses $\mathcal{C} = \{(Cl_i,C_i)\}_{i=1}^n$, where $Cl_i$ is a disjunction of literals of the form $P(\mathbf{X})$ or $\neg P(\mathbf{X})$, and $C_i$ is a constraint on the logvars. We transform this model to a parfactor model $M'$ consisting of two groups of parfactors: 
\begin{description}
\item [Weight parfactors] First we consider the weight function $w$. For each predicate $P$ in $M$ we add a parfactor $\phi_P(P(\mathbf{X}))$ to $M'$, with potential $\phi_P$ defined as: 
 $\phi_P(true) = w(P)$ and $\phi_P(false) = 1 - w(P)$.
 \item [Clause parfactors] Now we consider the set of constrained clauses $\mathcal{C}$. For each constrained clause $(Cl_i,C_i) \in \mathcal{C}$, we add a parfactor $\phi_i(\mathcal{A}_i) | C_i$ to $M'$, where $\mathcal{A}_i$ is the set of atoms that appear (in negated form) in clause $Cl_i$, and the potential $\phi_i$ is defined such that for any assignment of values $\mathbf{a}$ to $\mathcal{A}_i$: $\phi_i(\mathbf{a}) = 1$  if $\mathbf{a}$ satisfies $Cl_i$, and $\phi_i(\mathbf{a}) =0$ otherwise.
\end{description}
 
This transformation maps any WMC model $M$ to a parfactor model $M'$ that defines the same probability distribution as $M$. The following example illustrates such a transformation.

\textbf{Example.} Consider the 2-logvar WMC model $M$ consisting of the weight function $w$, and the constrained clause, $$\neg P(X) \vee Q(Y) | X \neq Y$$ 
Using the above method we derive the equivalent parfactor model $M'$ consisting of the following set of parfactors:
\begin{itemize} 
\item Two \emph{weight} parfactors $\phi_P(P(X))$ and $\phi_Q(Q(X))$, with potentials $\phi_P$ and $\phi_Q$ defined as follows:

\begin{center}
\begin{tabular}{c|c}
$P$ & $\phi_P$ \\
\hline
$false$ &$1 - w(P)$\\
$true$&$w(P)$\\
 \end{tabular} 
 \hspace{2 cm}
 \begin{tabular}{c|c}
$Q$ & $\phi_Q$ \\
\hline
$false$ &$1 - w(Q)$\\
$true$&$w(Q)$\\
 \end{tabular}
\end{center}

 \item One \emph{clause} parfactor $\phi(P(X),Q(Y)) | X \neq Y$, with potential function $\phi$ defined as follows:
\begin{center}
\begin{tabular}{cc|c}
$P$ & $Q$ & $\phi$ \\
\hline 
$false$ & $false$ & $1$\\
$false$ & $true$ & $1$\\
$true$ & $false$ & $0$\\
$true$ & $true$ & $1$\\
 \end{tabular} 
\end{center}
\end{itemize}
Note that the parfactor model $M'$, similar to the WMC model $M$, is a 2-logvar model. \exEnd

Given any WMC model $M$, this transformation maps each clause in $M$ to a parfactor that involves the same atoms, in the resulting parfactor model $M'$. As such, each clause is mapped to a parfactor with the same (number of) logvars. This transformation thus maps any $k$-WFOMC model into an equivalent $k$-logvar parfactor model.

%% file: fove_complete.bbl
\begin{thebibliography}{10}

\bibitem{Apsel2011}
Udi Apsel and Ronen~I. Brafman.
\newblock Extended lifted inference with joint formulas.
\newblock In {\em Proceedings of the 27th Conference on Uncertainty in
  Artificial Intelligence (UAI-11)}, pages 11--18. AUAI Press, 2011.

\bibitem{Darwiche01}
Adnan Darwiche.
\newblock Recursive conditioning.
\newblock {\em Artif. Intell.}, 126(1-2):5--41, 2001.

\bibitem{DeRaedt2008-PILP}
Luc De~Raedt, Paolo Frasconi, Kristian Kersting, and Stephen Muggleton,
  editors.
\newblock {\em Probabilistic inductive logic programming: theory and
  applications}.
\newblock Springer-Verlag, Berlin, Heidelberg, 2008.

\bibitem{deSalvoBraz2007}
Rodrigo {de Salvo Braz}.
\newblock {\em {Lifted first-order probabilistic inference}}.
\newblock PhD thesis, Department of Computer Science, University of Illinois at
  Urbana-Champaign, 2007.

\bibitem{Braz2005IJCAI}
Rodrigo {de Salvo Braz}, Eyal Amir, and Dan Roth.
\newblock Lifted first-order probabilistic inference.
\newblock In {\em Proceedings of the 19th International Joint Conference on
  Artificial Intelligence (IJCAI-05)}, pages 1319--1325, 2005.

\bibitem{Dechter99}
Rina Dechter.
\newblock Bucket elimination: A unifying framework for reasoning.
\newblock {\em Artif. Intell.}, 113(1-2):41--85, 1999.

\bibitem{Dechter07}
Rina Dechter and Robert Mateescu.
\newblock And/or search spaces for graphical models.
\newblock {\em Artif. Intell.}, 171(2-3):73--106, 2007.

\bibitem{GuyNips11}
Guy~Van den Broeck.
\newblock On the completeness of first-order knowledge compilation for lifted
  probabilistic inference.
\newblock In John Shawe-Taylor, Richard~S. Zemel, Peter~L. Bartlett, Fernando
  C.~N. Pereira, and Kilian~Q. Weinberger, editors, {\em Advances in Neural
  Information Processing Systems 24 (NIPS-11)}, pages 1386--1394, 2011.

\bibitem{Getoor07:book}
Lise Getoor and Ben Taskar, editors.
\newblock {\em An Introduction to Statistical Relational Learning}.
\newblock {MIT} {P}ress, 2007.

\bibitem{Gogate2011}
Vibhav Gogate and Pedro Domingos.
\newblock Probabilistic theorem proving.
\newblock In {\em Proceedings of the 27th Conference on Uncertainty in
  Artificial Intelligence (UAI-11)}, pages 247--256. AUAI Press, 2011.

\bibitem{Jha2010}
Abhay Jha, Vibhav Gogate, Alexandra Meliou, and Dan Suciu.
\newblock Lifted inference seen from the other side : The tractable features.
\newblock In J.~Lafferty, C.~K.~I. Williams, J.~Shawe-Taylor, R.S. Zemel, and
  A.~Culotta, editors, {\em Advances in Neural Information Processing Systems
  23 (NIPS-11)}, pages 973--981. 2010.

\bibitem{kersting09uai}
Kristian Kersting, Babak Ahmadi, and Sriraam Natarajan.
\newblock Counting belief propagation.
\newblock In {\em Proceedings of the 25th Conference on Uncertainty in
  Artificial Intelligence (UAI{--}09)}, 2009.

\bibitem{Kisynski2009IJCAI}
Jacek Kisynski and David Poole.
\newblock Lifted aggregation in directed first-order probabilistic models.
\newblock In {\em Proceedings of the 21th International Joint Conference on
  Artificial Intelligence (IJCAI-09)}, 2009.

\bibitem{Milch2008}
Brian Milch, Luke~S. Zettlemoyer, Kristian Kersting, Michael Haimes, and
  Leslie~Pack Kaelbling.
\newblock Lifted probabilistic inference with counting formulas.
\newblock In {\em Proceedings of the 23rd AAAI Conference on Artificial
  Intelligence (AAAI-08)}, pages 1062--1608, 2008.

\bibitem{Poole2003}
David Poole.
\newblock First-order probabilistic inference.
\newblock In {\em Proceedings of the 18th International Joint Conference on
  Artificial Intelligence (IJCAI-03)}, pages 985--991, 2003.

\bibitem{Poole2011}
David Poole, Fahiem Bacchus, and Jacek Kisynski.
\newblock Towards completely lifted search-based probabilistic inference.
\newblock {\em CoRR}, abs/1107.4035, 2011.

\bibitem{PooleZ03}
David Poole and Nevin~Lianwen Zhang.
\newblock Exploiting contextual independence in probabilistic inference.
\newblock {\em J. Artif. Intell. Res. (JAIR)}, 18:263--313, 2003.

\bibitem{Sen2009b}
Prithviraj Sen, Amol Deshpande, and Lise Getoor.
\newblock Bisimulation-based approximate lifted inference.
\newblock In A.~Ng and J.~Bilmes, editors, {\em Proceedings of the 25th
  Conference on Uncertainty in Artificial Intelligence (UAI{--}09)}, June
  18{--}21 2009.

\bibitem{Singla2008}
Parag Singla and Pedro Domingos.
\newblock Lifted first-order belief propagation.
\newblock In {\em Proceedings of the 23rd AAAI Conference on Artificial
  Intelligence (AAAI-08)}, pages 1094--1099, 2008.

\bibitem{Taghipour2012}
Nima Taghipour, Daan Fierens, Jesse Davis, and Hendrik Blockeel.
\newblock Lifted variable elimination with arbitrary constraints.
\newblock In {\em Proceedings of the 15th International Conference on
  Artificial Intelligence and Statistics (AISTATS-12)}. JMLR Workshop and
  Conference Proceedings, 2012.

\bibitem{GuyIJCAI11}
Guy Van~den Broeck, Nima Taghipour, Wannes Meert, Jesse Davis, and Luc
  De~Raedt.
\newblock Lifted {P}robabilistic {I}nference by {F}irst-{O}rder {K}nowledge
  {C}ompilation.
\newblock In Toby Walsh, editor, {\em Proceedings of the 20th International
  Joint Conference on Artificial Intelligence (IJCAI-11)}, July 2011.

\end{thebibliography}
